\documentclass[journal,onecolumn]{IEEEtran}
\ifCLASSINFOpdf
\else
\fi

\usepackage{amssymb}
\usepackage{amsmath}
\usepackage{graphicx}
\usepackage{subfigure}
\usepackage{multirow}
\usepackage{booktabs}
\usepackage{color}
\usepackage[ruled,linesnumbered]{algorithm2e}
\usepackage{bm}

\def\scx{\textsc{x}}

\graphicspath{{fig/}}

\begin{document}

\title{Weakly supervised segment annotation via expectation kernel density estimation}

\author{{Liantao~Wang, Qingwu Li, Jianfeng Lu} %
\thanks{Liantao~Wang and Qingwu Li are with the College of Internet of Things Engineering, Hohai University, Changzhou 213022, China e-mail: (ltwang@hhu.edu.cn).
}
\thanks{Jianfeng Lu is with the Jiangsu Key Laboratory of Image and Video Understanding for Social Safety, Nanjing University of Science and Technology, Nanjing 210094, China}
}

%



\maketitle

%
\begin{abstract}
Since the labelling for the positive images/videos is ambiguous in weakly supervised segment annotation, negative mining based methods that only use the intra-class information emerge. In these methods, negative instances are utilized to penalize unknown instances to rank their likelihood of being an object, which can be considered as a voting in terms of similarity. However, these methods 1) ignore the information contained in positive bags; 2) only rank the likelihood but cannot generate an explicit decision function. In this paper, we propose a voting scheme involving not only the definite negative instances but also the ambiguous positive instances to make use of the extra useful information in the weakly labelled positive bags. In the scheme, each instance votes for its label with a magnitude arising from the similarity, and the ambiguous positive instances are assigned soft labels that are iteratively updated during the voting. It overcomes the limitations of voting using only the negative bags. We also propose an expectation kernel density estimation (eKDE) algorithm to gain further insight into the voting mechanism. Experimental results demonstrate the superiority of our scheme beyond the baselines.
\end{abstract}
%

%
\IEEEpeerreviewmaketitle

\section{Introduction}
With the development of communication technology and the popularity of digital cameras, one can easily access massive images/videos. Although these digital multimedia are usually associated with semantic tags indicating certain visual concepts appearing inside, the exact locations remain unknown, leading to their infeasibility for training traditional supervised visual recognition models. As a result, there has been a great interest in object localization for images/videos with weak labels \cite{Nguyen09, Siva12, ZhouJP15,Bilen16,HanQZN18}.

An alternative is weakly supervised segment annotation (WSSA) \cite{Viola05, Li09, Tang13, Zhao16}. For images/videos with weak labels, those with objects of interest inside are considered as positive bags, while those without objects of interest are negative. Based on unsupervised over-segmentation, images/videos are transformed into segments, and the task is to distinguish whether they correspond to a given visual concept.

Among the state-of-the-art methods for weakly supervised segment annotation (WSSA), there is a simple yet effective branch \cite{Siva12,LiY09,Tang13,KrapacPFJ14}. They employ the inter-class or intra-class information by measuring similarities between instances based on two rules: 1) Positive instances are similar patterns existing in different positive bags; 2) Positive instances are dissimilar to all the instances in negative bags. For an unknown instance, they iterates through the labelled instances, and each gives a vote for or against its being a target.

In \cite{Siva12,Tang13,Jiang15a,WangLLW18}, the authors insist that inter-class information is more useful in a MIL setting, and propose to use negative instances to vote against the unknown instances, and select that least penalized as the instance of interest. In these methods, only negative instances with definite labels are eligible to vote. It is true that the number of negative instances is much larger than that of potential positive instances, and the labels are also more definite. However useful information in positive bags is ignored. However, there are two limitations for these methods. 1) Useful information in positive bags is ignored. 2) Only a ranking of likelihood instead of an explicit decision function is output. Although thresholding the ranking can generate a classification, there is not a strategy to theoretically decide the threshold value.

In this paper, we argue that extra useful information can be mined from the weakly labelled positive bags besides the definite negative bags. Consequently the instances can be annotated by looking at the weakly labelled data themselves. Therefore we proposed a self-voting scheme, where all the instances are involved. The contributions of this paper are as follows:
1) A voting scheme involving both negative instances and ambiguous instances in positive bags is proposed.
2) The proposed voting scheme can output discriminant results beyond just ranking.
3) An expectation kernel density estimation (eKDE) algorithm is proposed to handle weakly labelled data. A deep interpretation is provided from the maximum posterior criterion (MAP) and eKDE for the proposed voting scheme
4) Relations to existing methods including negative mining, supervised KDE and semi-supervised KDE, are analyzed.

In a WSSA task, two sets of images (the same for videos) are given with image-level labels. Each image in the positive set contains an instance of an identical object category, and each image in the negative set does not contain any instance of the object category. Negative mining methods determine a region in a positive image the likelihood of being an object of interest by its dissimilarity to the negative regions. Besides this inter-class information, our method further takes into account the intra-class information that all the object regions in different positive images should have high similarity because they come from an identical object category.The extra information improves the performance compared to negative mining.

The remainder of this paper is organized as follows. Section \ref{sec_related} reviews the related works. We then detail the methodology in Section \ref{sec_method}.  We first revisit the negative mining methods in a voting framework (\ref{sec_revisit}), then propose our weighted self-voting scheme (\ref{sec_scheme}). To get an insight into the mechanism of our scheme, we derive an interpretation from MAP and eKDE (\ref{sec_interpretation}). Difference from other existing methods are also analysed (\ref{sec_diff}).  Experimental results are reported in Section \ref{sec_exp} . Section \ref{sec_conc} concludes this work.

\section{Related work}
\label{sec_related}
Negative mining methods train a classifier based on the strongly labelled negative training data. For each instance in a positive bag, based on the inter-class information, NegMin \cite{Siva12} compute their similarities with all of the negative instances, and select the instance that has minimum max-similarity as of interest. CRANE \cite{Tang13} selects negative instances to vote against an unknown instance by specifying some similarity threshold, and improves the robustness of labelling noise among negative instances. Fu et al. \cite{Fu11} also make use of the similarity information as a pre-processing heuristic for a bag-level classification. They select instances with least similarity to the negative bags and use them to initialize cluster centers, which are then used to create the bag level feature descriptors of \cite{Chen06}. Moreover, Jiang \cite{Jiang15a} trains a one-class SVM based on negative instances, then ranks the saliency according to the distances to the decision boundary.

Besides using the inter-class information, key instance detection can be accomplished by searching similar patterns among diverse positive bags. The most classical framework is diverse density (DD) \cite{Maron97}. It defines a conditional probability with similarity, and uses the noisy-or model to define a diverse density to select instances with high similarities to diverse positive bags and low similarities to negative bags. DD has been widely used as a basis for many methods including EM-DD\cite{Zhang01}, GEM-DD \cite{Rahmani08} and DD-SVM \cite{ChenW04}. However, DD is sensitive to labelling noise. Evidence confidence \cite{LiY09} is proposed to seek the mode on observed instances rather than in a continuous space to facilitate the computation and alleviate the sensitivity. Krapac et al. \cite{KrapacPFJ14} exploit similarity among class-specific features to decide prototypes, which are used in a voting-based mechanism to select instances with a high diverse occurrence.

Since we derive a KDE interpretation for our voting scheme, we also make a literature review on this subject. KDE possesses the advantages of nonparametric method for unsupervised density estimation. Du et al. \cite{DuWHY13} propose a supervised KDE to make use of labels, and extend the mean shift \cite{Comaniciu02} to a supervised version to seek modes. In order to make full use of unlabelled data, Wang et al. \cite{WangSYZHL06,WangHMHQSD09} propose a semi-supervised KDE to estimate class-specific density based on a little fraction of labelled data. SSKDE is later extended to a manifold structure \cite{JiZHJ14}.

Shallow learning methods have been outperformed by deep convolutional neural networks (DCNN) significantly on the visual recognition tasks resulting from their powerful feature representation \cite{ChatfieldSVZ14}. One approach to boosting the performance of shallow methods is using deep features from pre-trained DCNN models. R-CNN \cite{GirshickDDM14} combined SVM with DCNN features to boost the object detection performance. DCNN features have also been incorporated into weakly supervised visual recognition tasks. Zhang et al. \cite{ZhangMH17} concatenate multiple convolutional outputs and max-pool them to represent the super-pixel features. Observing that a region probably belongs to an object if many channels of the hidden-layer activation fire simultaneously, Wei et al. \cite{WeiLWZ17} select the object regions using aggregation map, then max-pool the concatenation of multiple-layer activations to represent the image. Similarly, based on the findings that the hidden-layer activations of a pre-trained object recognition network usually fire up on objects rather than background,  Saleh et al. \cite{SalehASPAG18} leverage these masks for weakly supervised semantic segmentation.

\section{Methodology}
\label{sec_method}
In a weakly supervised learning scenario, a label is given at a coarser level, and accounts for a collection of instances rather than for individual instance, usually for the purpose of efforts reduction. A positive label indicates that the collection contains at least one instance of interest, while a negative one indicates that none of the collection is of interest. Such data can be naturally represented by bags that arise from multiple-instance learning. Without loss of generality, we denote such data by $\mathcal{D}  = \{ \langle \mathcal{B}_i, y_i \rangle \}_{i=1}^{m}$, where $\mathcal{B}_i = \{ \scx_{ij} \}_{j=1}^{|\mathcal{B}_i|}$ is a bag, with $\scx_{ij} \in \mathbb{R}^D$ an instance and $y_i \in \{1,-1 \}$ a label. The data annotation is to predict $y_{ij} \in \{1,-1 \}$ for each instance. For the sake of clarity, we separate  notations of positive bags and negative bags, then the sample set $\mathcal{D} = \{ \langle \mathcal{B}_i^+ \rangle \}_{i=1}^{p} \bigcup \{ \langle \mathcal{B}_i^- \rangle \}_{i=1}^{n}$, where we assume that the numbers of positive bags and  negative bags are $p$ and $n=m-p$, respectively.

\subsection{Negative mining revisited}
\label{sec_revisit}
Negative mining methods \cite{Siva12, Tang13} insist that, in the scenario of WSSA, the much larger amount of negative instances provide more useful information. Therefore they only make use of the negative bags with definite labels, and ignore the ambiguous information of positive bags, to localize objects of interest.
For a given positive bag, NegMin \cite{Siva12} selects the instance that minimize the similarity to the nearest neighbour in the collection of the negative instances. Let $s_{ij}\triangleq s(\scx,\scx_{ij}) >0$ denotes the similarity of $\scx$ and $\scx_{ij}$. The notion of NegMin can be formalized as follows. It scores an instance by
\begin{equation}\label{eq_negmin}
f_{NegMin} (\scx) = \min \sum_{i=1}^{n} \sum_{j}^{|\mathcal{B}_i^-|} - u_{ij} \cdot s_{ij}
\end{equation}
with $u_{ij} \in \{0,1\},~ \sum_{j=1}^{|\mathcal{B}_i^-|} u_{ij} =1~ \forall i$. Then the  $j^*$-th instance with the maximum score in a positive bag $\mathcal{B}_i^+$ is considered as the instance of interest:
\begin{equation}
j^* = \arg\max _{j \in \{1, \cdots, |\mathcal{B}_i^+| \}} f_{NegMin}(\scx_{ij}), ~ i=1, \cdots, p.
\end{equation}

Similarly, from the negative mining perspective, CRANE \cite{Tang13} selects instances from the negative bags to penalize their nearby instances in the positive bags, by the following scoring strategy:
\begin{equation}\label{eq_crane}
f_{CRANE}(\scx) = \sum_{i=1}^{n} \sum_{j}^{|\mathcal{B}_i^-|} -s_{cut}(s_{ij}) \cdot \delta(s_{ij} \not< \Delta).
\end{equation}
A naive constant $s_{cut}(\cdot) =1$ is used in \cite{Tang13}. $\delta(\cdot)$ denotes the indicator function, and $\Delta = \max _{t} s(\scx_{ij},\scx_{t})$ makes only the negative instances, which have $\scx$ as its nearest neighbour in the positive bags, can vote a penalty. In the ambiguous positive bags, negative instances are usually similar to those in negative bags, while the concept instances are rarely the closest to negative instances. As a result, negative instances will be more penalized, and scored lower than those potential concept instances.

For both \eqref{eq_negmin} and \eqref{eq_crane}, the instance scored higher are more likely a concept instance. For NegMin \cite{Siva12}, the instance with the maximum score is considered as the object, which makes it infeasible for multiple instance detection \cite{Liu12}. Although CRANE \cite{Tang13} is able to rank the likelihood of the instances being of interest, there is not an explicit classification boundary, therefore a threshold is needed to manually set to generate concept instances. Moreover, these methods are usually sensitive to outliers, since they only employ the instances with extreme similarities for voting.

\subsection{Weighted self-voting}
\label{sec_scheme}
In order to address the above limitations, we seek a voting scheme using both inter-class information and underlying intra-class information of positive instances. Suppose we already have instances with definite labels, to develop a reasonable voting scheme, each instance should vote to an unknown $\scx$ for the label of itself according to their similarity, i.e., for a more similar instance, its voting magnitude should be larger, and vice versa. We then weight the voting by similarity, and yield a voting term of $\scx_{ij}$ with a label $y_{ij}$ for $\scx$:
\begin{equation}\label{eq_labVot}
f_{ij}(\scx) = y_{ij} \cdot s_{ij}  .
\end{equation}

For the case of weakly labelled data, the labels for some instances are ambiguous. We therefore introduce another weight $w_{ij} \in [0,1]$ to denote the likelihood of $\scx_{ij}$ having a positive label, and change the voting term to:
\begin{align}\label{eq_votingterm}
f_{ij}(\scx) = \left\{
\begin{array}{ll}
w_{ij} \cdot 1 \cdot s_{ij}   & \text{for }y_{ij}=1; \\
(1-w_{ij}) \cdot (-1) \cdot s_{ij}    & \text{for } y_{ij}=-1.
\end{array} \right.
\end{align}
In other words, $w_{ij}\triangleq p(y_{ij}=1|\scx_{ij})$.

Then given a set of weakly labelled bags, we can obtain the voting score for an unknown instance $\scx$ as follows:
\begin{align}\label{eq_votingscheme}
f(\scx) = \sum \limits_{i=1}^{p} \sum \limits_{j}^{|\mathcal{B}_i^+|}  w_{ij} s_{ij}  - \Big( \sum \limits_{i=1}^{n} \sum \limits_{j}^{|\mathcal{B}_i^-|} s_{ij} + \sum \limits_{i=1}^{p} \sum \limits_{j}^{|\mathcal{B}_i^+|} (1-w_{ij})  s_{ij} \Big),
\end{align}
where we can see that each instance votes with a magnitude $s_{ij}$ for the label of itself that is definite or ambiguous. A negative instance votes for a definite $-1$, and $w_{ij}$  can be considered as a soft label that is introduced for ambiguous labels.

Here we intuitively explain why employing ambiguous instances to vote is reasonable. More formal interpretation from the viewpoints of MAP and eKDE can be seen in later section. For an instance $\scx$, each instance votes for its own label with a value measuring their similarity \eqref{eq_labVot} and \eqref{eq_votingscheme}.
A potential object instance has many strong supporters existing in each positive image, because all the positive images contain same class of objects.
In other words, among all of the votes, those positive values from its supporters are large due to high intra-class similarities, and the negative votes from its protesters are small due to low inter-class similarities.
While a potential negative instance does not have many supporters because this pattern does not appear in all of the positive images, and all of the positive votes tend to be small. By contrast, it is more possible to be similar to the background, and obtain high negative vote values that will suppress the small positive ones.

We expect \eqref{eq_votingscheme} to be able to generate an explicit label for instance $\scx$ by:
\begin{equation}\label{eq_sgn}
\hat{y} = sgn(f(\scx)).
\end{equation}
Intuitively for an segment, when the voting for its being an object overwhelms that against its being an object, Eq. \eqref{eq_votingscheme} gives a positive value to classify it as an object, and vice versa. Later in this paper, we will demonstrate that \eqref{eq_votingscheme} actually complies with the MAP criterion under a expectation Kernel Density Estimation algorithm.

Note that our voting scheme \eqref{eq_votingscheme} makes use of the ambiguous positive bags as well as the definite negative instances. Both NegMin and CRANE are special cases of the formulation \eqref{eq_votingscheme} that only involves negative instances. If we only use the negative instances with definite labels, \eqref{eq_votingscheme} becomes:
\begin{equation}\label{eq_negvoting}
f_{neg}(\scx)  =  - \sum \limits_{i=1}^{n} \sum \limits_{j}^{|\mathcal{B}_i^-|} s_{ij} ,
\end{equation}
which is actually the voting aggregation of all the negative instances, and is a reduced version of \cite{WangLLW18}. NegMin only picks the minimum of the voting, as seen in \eqref{eq_negmin}. CRANE selects part of the negative instances by $\delta$ to vote, and the voting magnitude is cut off by $f_{cut}$. Since NegMin and CRANE use instances with extreme similarities to vote, they are sensitive to outliers, while our voting scheme is much more robust by considering all of the sample. In addition, our scheme is able to mine the useful information contained in the ambiguous bags, and output the category of an instance.

\subsection{Interpretation from MAP and eKDE}
\label{sec_interpretation}
We interpret the scoring scheme \eqref{eq_votingscheme} and \eqref{eq_sgn} from the viewpoints of MAP and eKDE. Given an instance $\scx_{ij}$, we consider its label as a binary random variable $y_{ij} \in \{1,-1 \}$, where $1$ and $-1$  represent the positive class and the negative class respectively. Note that it is a Bernuulli distribution. When we describe the probability of $y_{ij}=1$ by the parameter $w_{ij}$, the probability distribution can be written in the form
\begin{equation}
p( y_{ij} | \scx_{ij} )  =  w_{ij} ^ {\frac{1+y_{ij}}{2}} (1-w_{ij}) ^ {\frac{1-y_{ij}}{2}}.
\end{equation}

Suppose we already have the labels $y_{ij}$ for instance $\scx_{ij}$ in each $\mathcal{B}_i^+$, and denote a kernel function by $k_{ij}\triangleq k(\scx,\scx_{ij})$, we can estimate the class conditional probabilities using the conventional KDE as follows:
\begin{align}
p^{*}(\scx | y=1) ~ = ~ & \frac{\sum \limits_{i=1}^{p} \sum \limits_{j}^{|\mathcal{B}_i^+|}  \frac{1+y_{ij}}{2} \cdot k_{ij} } {\sum \limits_{i=1}^{p} \sum \limits_{j}^{|\mathcal{B}_i^+|} \frac{1+y_{ij}}{2} },\label{eq_kde_pos}   \\
p^{*}(\scx | y=-1) ~ = ~ & \frac{\sum \limits_{i=1}^{n} \sum \limits_{j}^{|\mathcal{B}_i^-|} k_{ij} + \sum \limits_{i=1}^{p} \sum \limits_{j}^{|\mathcal{B}_i^+|} \frac{1-y_{ij}}{2} \cdot k_{ij} } {\sum \limits_{i=1}^{n} \sum \limits_{j}^{|\mathcal{B}_i^-|} 1 + \sum \limits_{i=1}^{p} \sum \limits_{j}^{|\mathcal{B}_i^+|} \frac{1-y_{ij}}{2} }. \label{eq_kde_neg}
\end{align}
In contrast to a fully conventional KDE instance, the difference is that $y_{ij}$ here are random variables rather than  constants. Consequently we have to compute the density using the expectation over the random variables $y_{ij}$:
\begin{align} \label{eq_ekde}
p(\scx | y=1) ~ = ~ & \mathbb{E}_{y_{ij}}  [p^*(\scx | y=1)] =  \frac{\sum \limits_{i=1}^{p} \sum \limits_{j}^{|\mathcal{B}_i^+|}  w_{ij} \cdot k_{ij} } {\sum \limits_{i=1}^{p} \sum \limits_{j}^{|\mathcal{B}_i^+|} w_{ij} },\nonumber    \\
p(\scx | y=-1) ~ = ~ & \mathbb{E}_{y_{ij}}  [p^*(\scx | y=-1)] \nonumber\\
= ~ & \frac{\sum \limits_{i=1}^{n} \sum \limits_{j}^{|\mathcal{B}_i^-|} k_{ij} + \sum \limits_{i=1}^{p} \sum \limits_{j}^{|\mathcal{B}_i^+|} (1-w_{ij}) \cdot k_{ij} } {\sum \limits_{i=1}^{n} \sum \limits_{j}^{|\mathcal{B}_i^-|} 1 + \sum \limits_{i=1}^{p} \sum \limits_{j}^{|\mathcal{B}_i^+|} (1-w_{ij}) }.
\end{align}
Eq. \eqref{eq_ekde} estimates probability density using kernel functions with expectation over extra random variables. We call them expectation kernel density estimation (eKDE).

Then the decision scheme \eqref{eq_votingscheme} and \eqref{eq_sgn} has an interpretation of MAP criterion. For an instance $\scx$, MAP decides its label by:
\begin{equation}\label{eq_max_post}
\hat{y} = \arg \max_{y \in \{-1, 1\} } p(y|\scx).
\end{equation}
From the Bayes' theorem, we have
\begin{equation}
p(y|\scx) \propto p(\scx | y) p(y).
\end{equation}
Then \eqref{eq_max_post} is equivalent to,
\begin{equation} \label{eq_sgnprobdiff}
\hat{y} = sgn(p(\scx | y=1) p(y=1) - p(\scx | y=-1) p(y=-1)).
\end{equation}

As a typical approach in machine learning, we can aggregate the posterior probabilities to approximate the effective number of points assigned to a class, and estimate the class priors $p(y)$ by the fractions of the data points assigned to each of the classes.
\begin{align}\label{eq_prior}
p(y=1) ~ = ~ & \sum \limits_{i=1}^{p} \sum \limits_{j}^{|\mathcal{B}_i^+|} w_{ij} / N,    \nonumber \\
p(y=-1) ~ = ~ & \Big( \sum \limits_{i=1}^{n} \sum \limits_{j}^{|\mathcal{B}_i^-|} 1 + \sum \limits_{i=1}^{p} \sum \limits_{j}^{|\mathcal{B}_i^+|} (1-w_{ij}) \Big) /N,
\end{align}
where $N$ denotes the total number of data points and can be omitted during computing the decision values. Using $k_{ij}$ measuring the similarity $s_{ij}$, and substituting \eqref{eq_ekde} and \eqref{eq_prior} into \eqref{eq_sgnprobdiff}, we obtain a discriminant function exactly the same as \eqref{eq_sgn}. Therefore our weighted voting scheme complies with the MAP criterion when using the proposed eKDE for weakly supervised density estimation.
\begin{figure*}[!t]
\centering
\begin{tabular}{cc}
  \vspace{+0.05in}
  \rotatebox{90}{\quad \quad \quad PittCar}&
  {\includegraphics[width=0.23\linewidth]{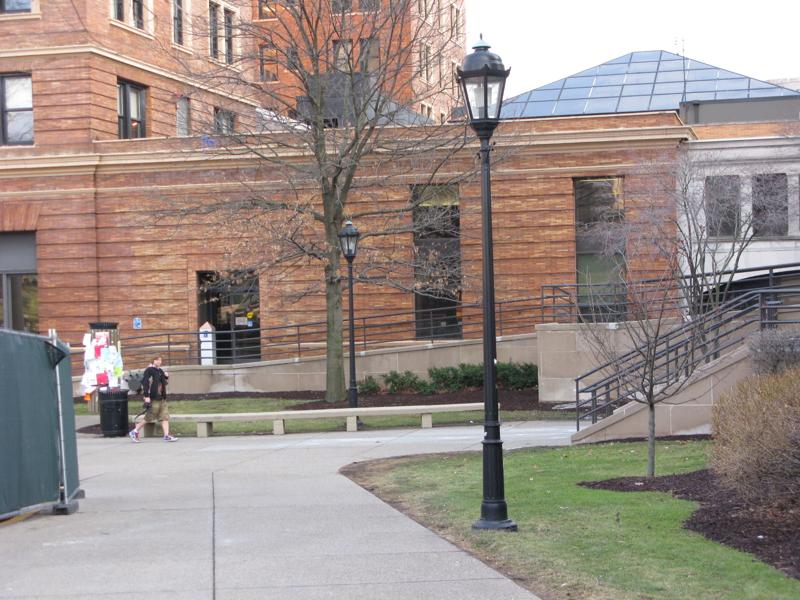}}
  {\includegraphics[width=0.23\linewidth]{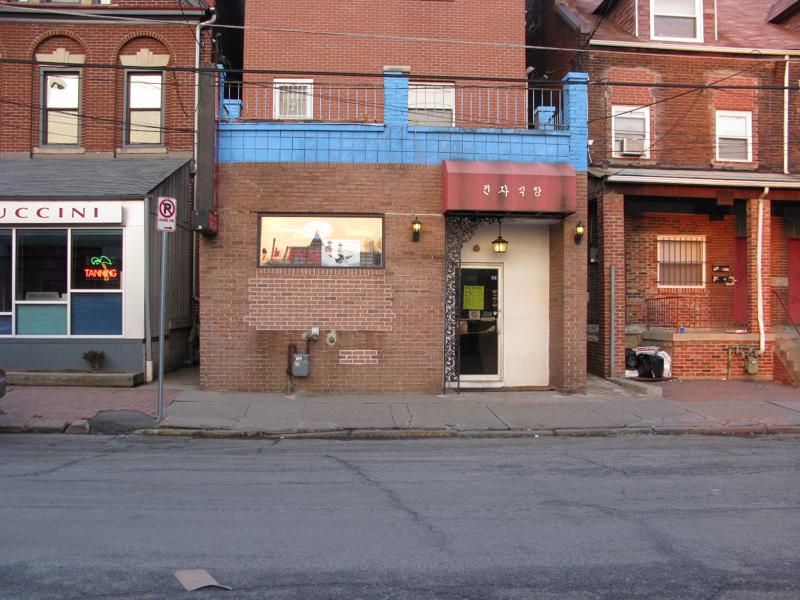}}
  {\includegraphics[width=0.23\linewidth]{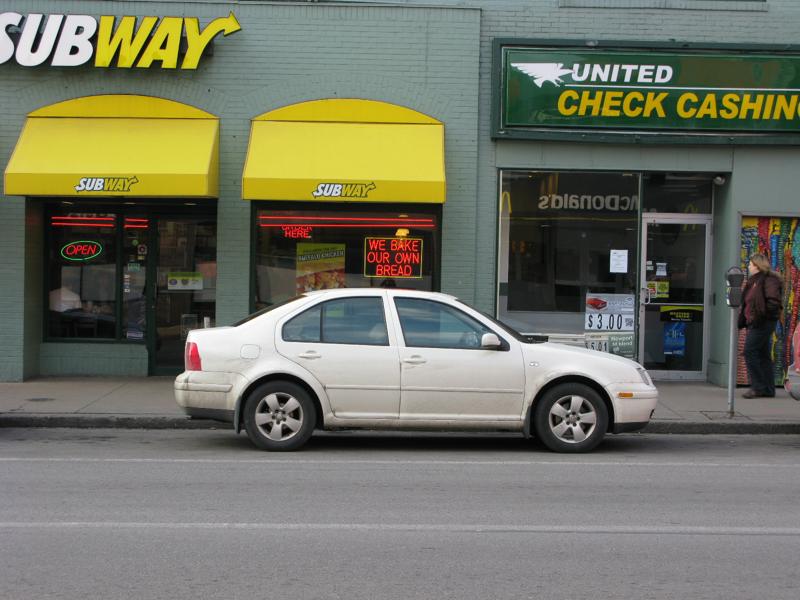}}
  {\includegraphics[width=0.23\linewidth]{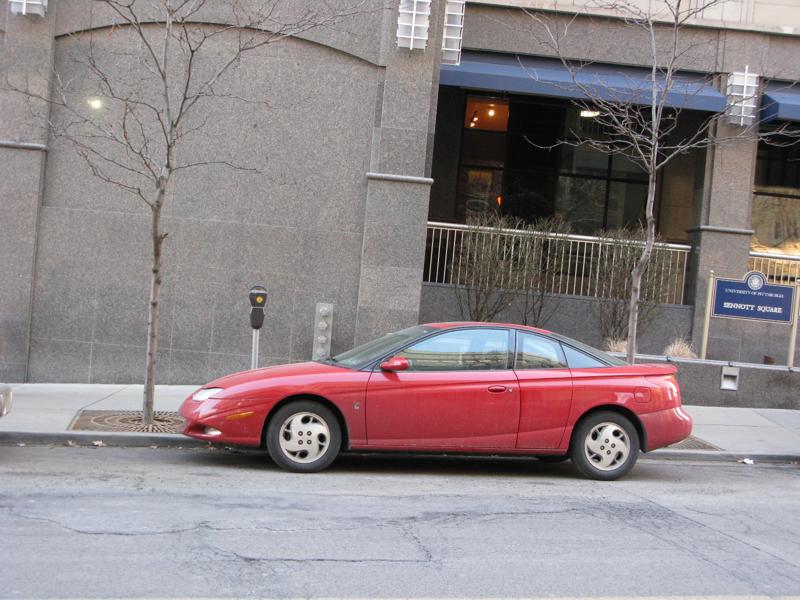}}  \\
  \vspace{+0.05in}
  \multirow{2}{*}{\rotatebox{90}{YTO}}&
  {\includegraphics[width=0.23\linewidth]{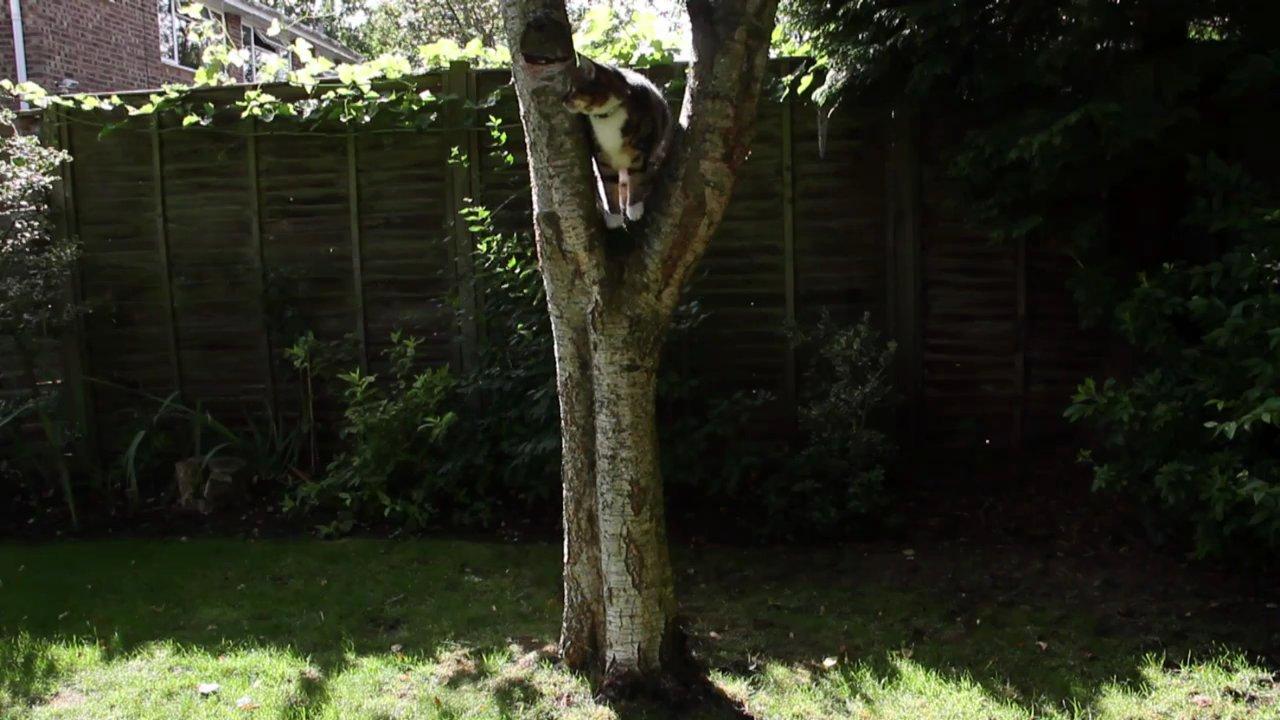}}
  {\includegraphics[width=0.23\linewidth]{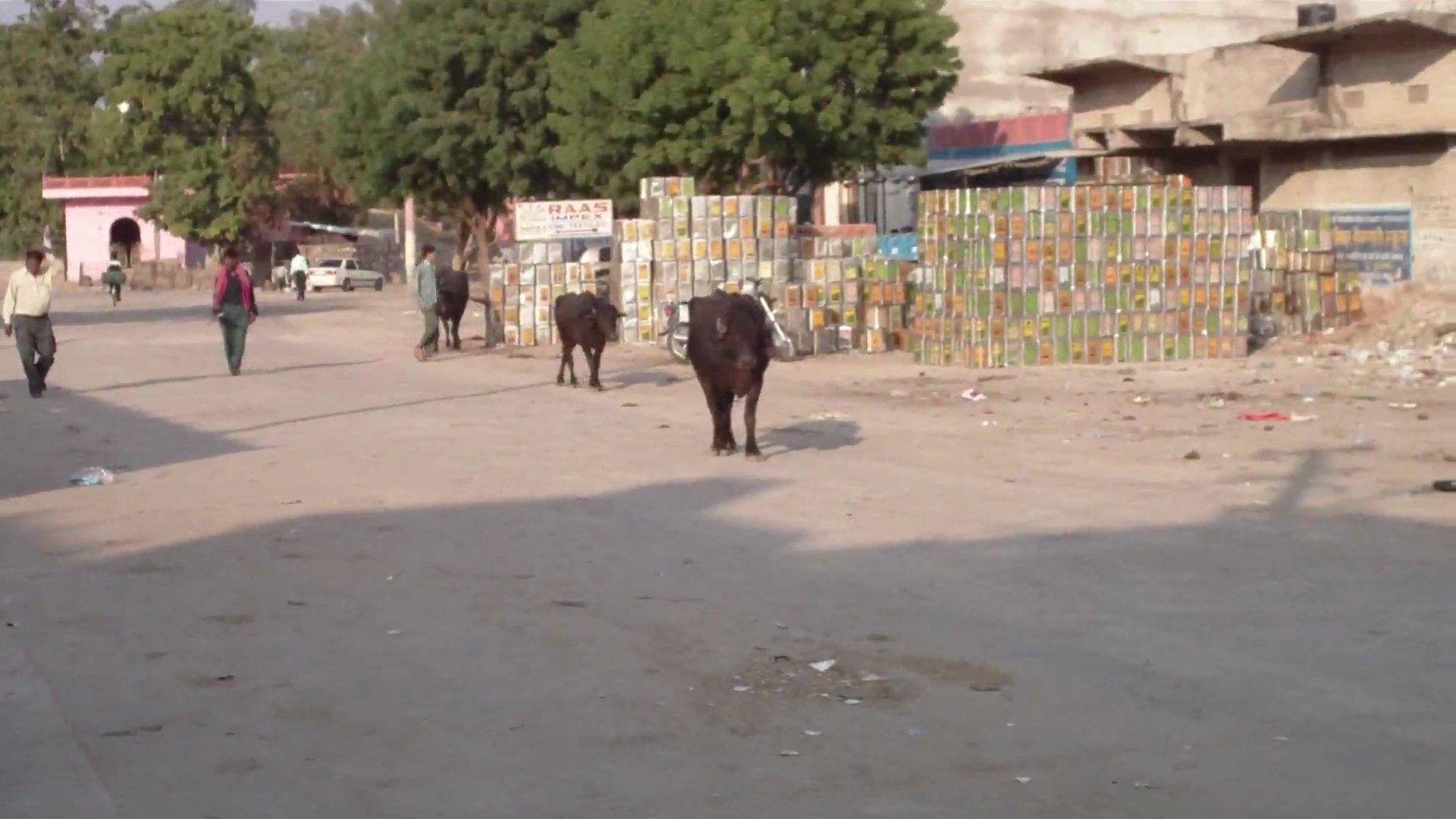}}
  {\includegraphics[width=0.23\linewidth]{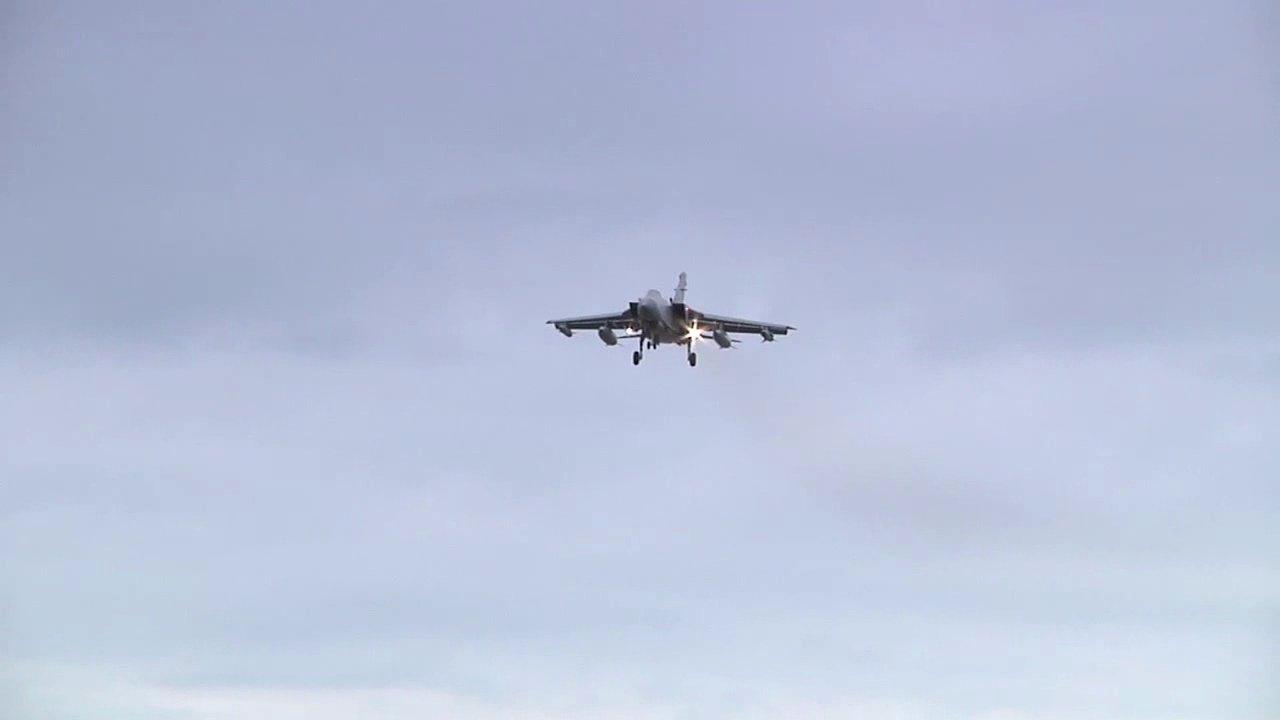}}
  {\includegraphics[width=0.23\linewidth]{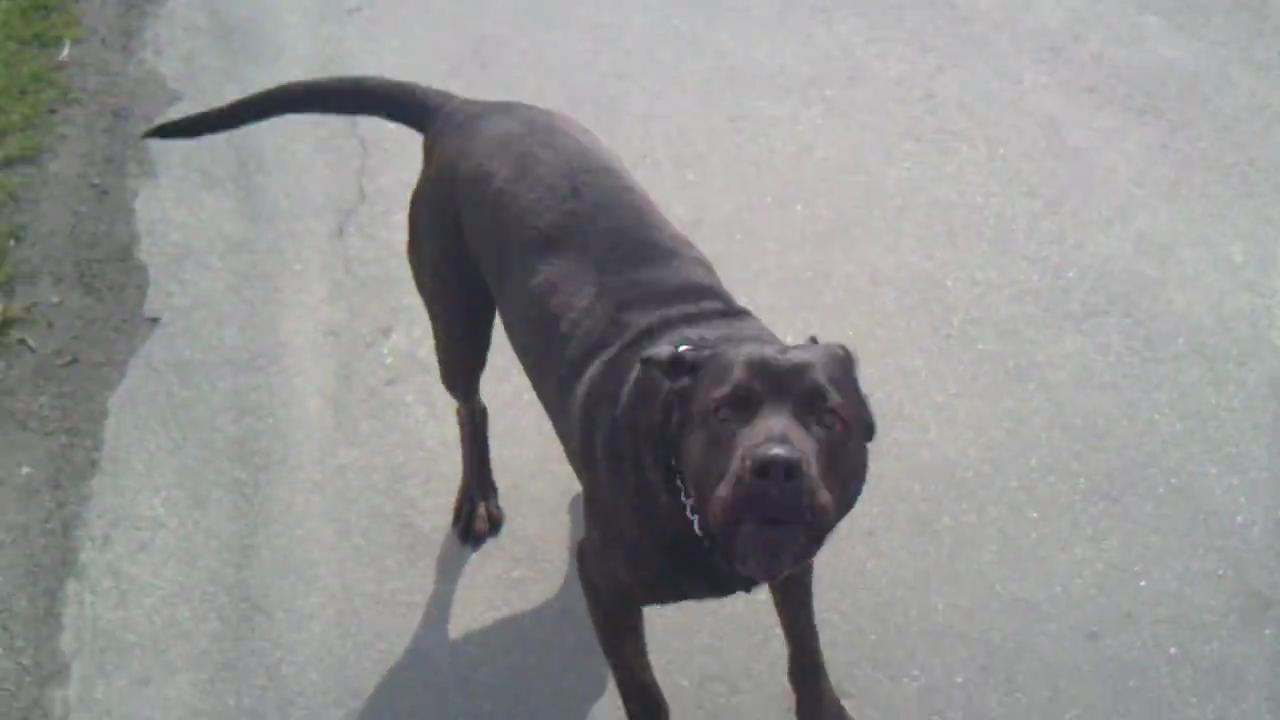}} \\
   &
  {\includegraphics[width=0.23\linewidth]{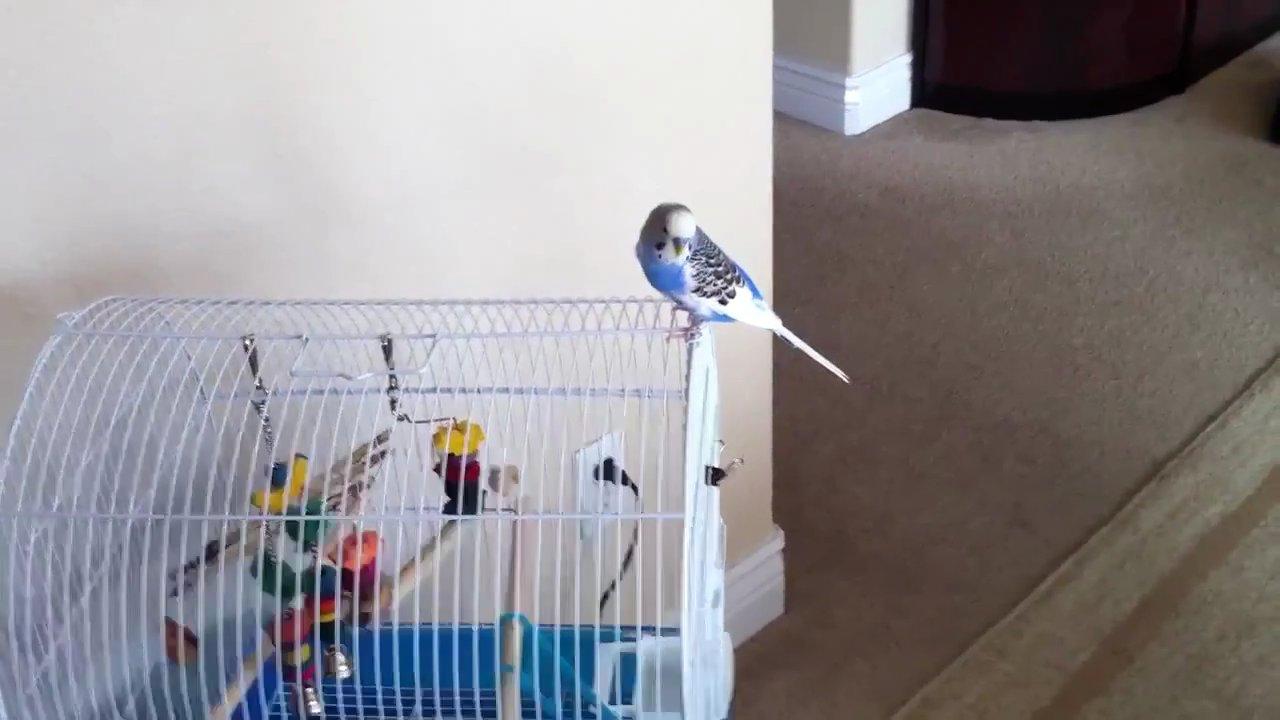}}
  {\includegraphics[width=0.23\linewidth]{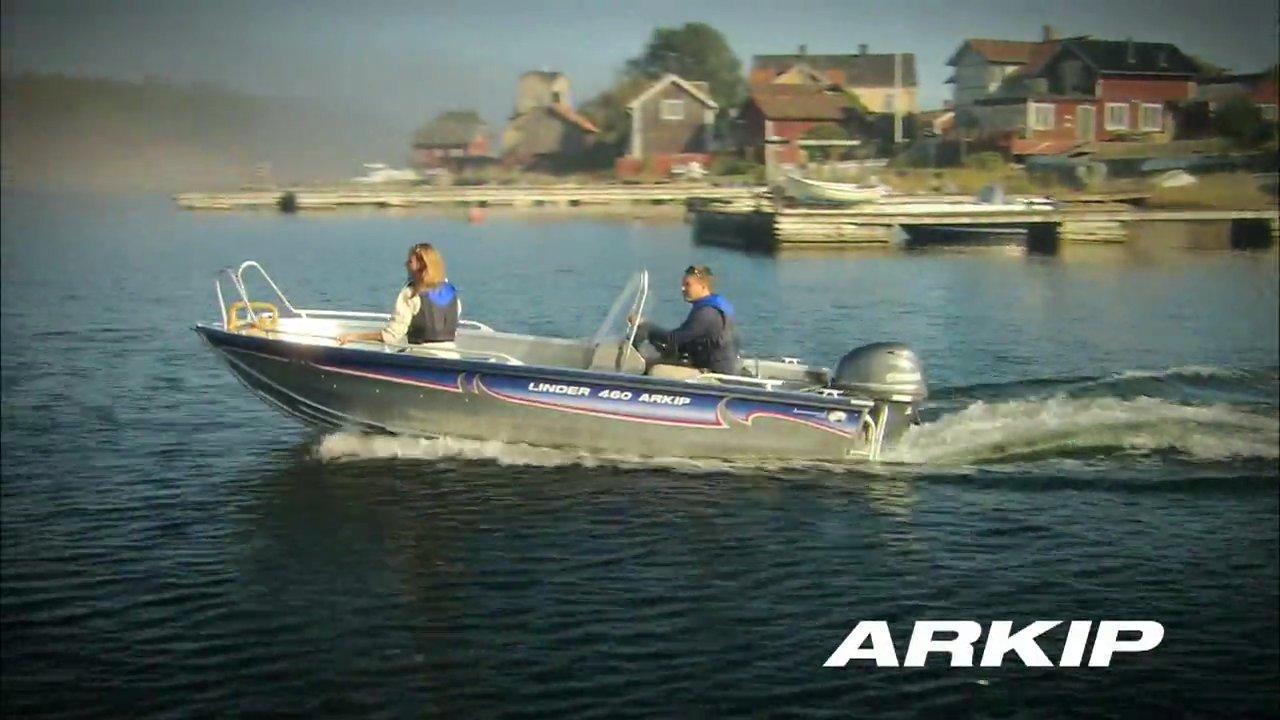}}
  {\includegraphics[width=0.23\linewidth]{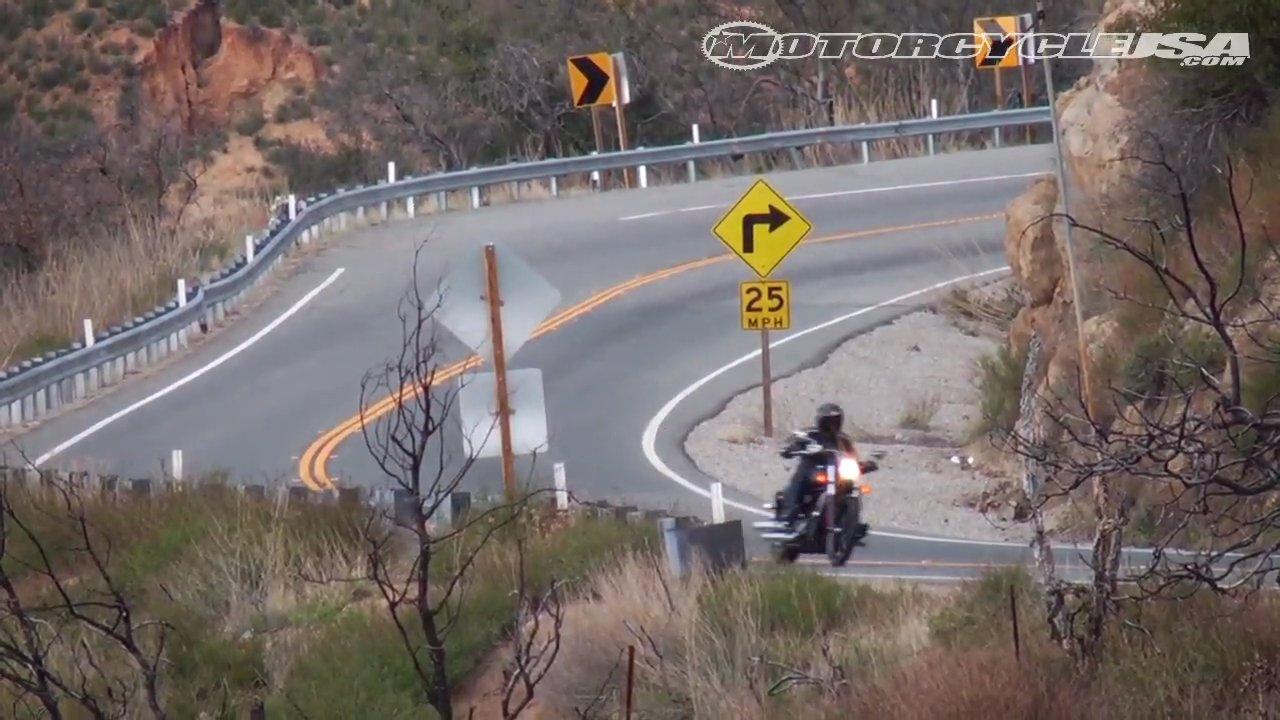}}
  {\includegraphics[width=0.23\linewidth]{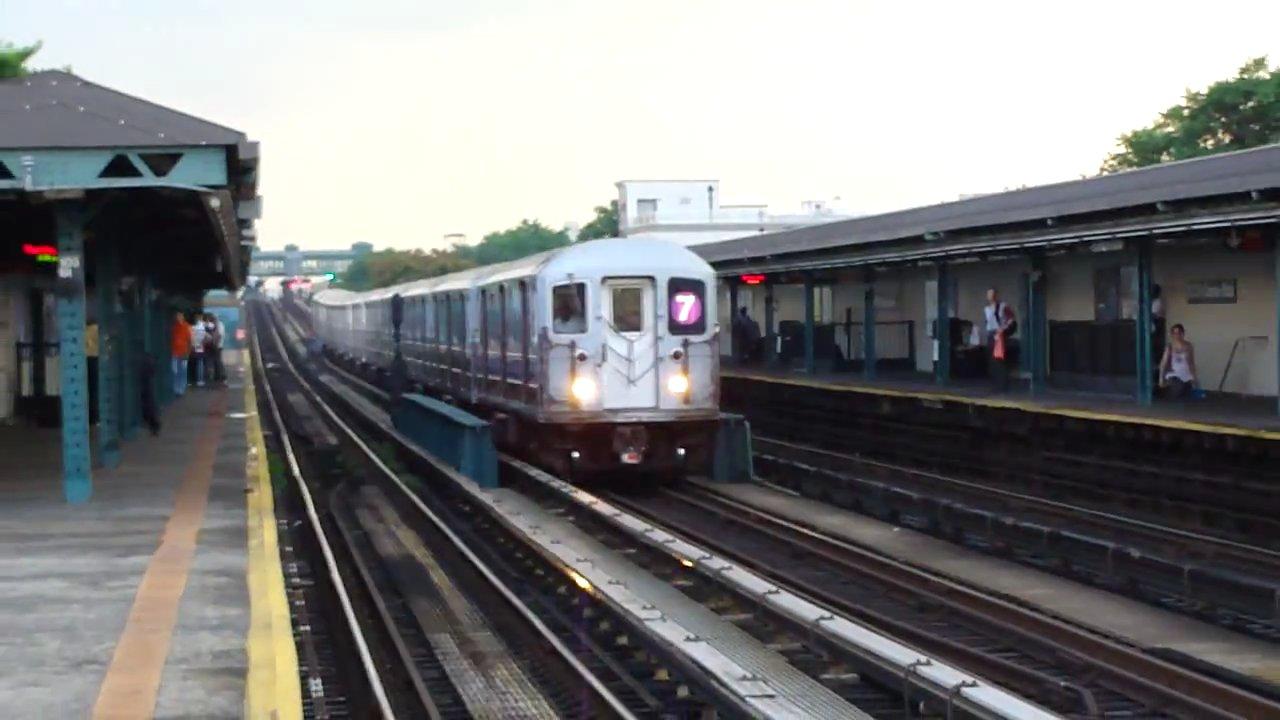}} \\
\end{tabular}
\caption{Some examples from datasets: PittCar (row 1), YTO (rows 2 and 3).}
\label{fig_data_example}
\end{figure*}

\subsection{Algorithm}
From the above demonstration, we can determine the label for a segment using Eqs. \eqref{eq_votingscheme} and \eqref{eq_sgn} equivalent to Eq. \eqref{eq_sgnprobdiff} that is a weighted difference of class conditional probability densities. On the one hand, the estimation of class conditional probability density $p(\scx|y)$ is dependent on the post probabilities $w_{ij}$ through \eqref{eq_ekde}. On the other hand, the post probabilities $w_{ij}$ are dependent on $p(\scx|y)$ through a simple deduction using the Bayes' theorem and the sum rule of probability:
\begin{align}\label{eq_w_update}
w_{ij} =  \frac{p(\scx_{ij} | y=1)p(y=1)}{p(\scx_{ij} | y=1)p(y=1)+p(\scx_{ij} | y=-1)p(y=-1)},
\end{align}
This mutual dependency naturally induces an iteratively method to solve the problem, which is described in Algorithm \ref{alg_1}.
\begin{algorithm}[t]
\caption{WSSA via eKDE.}
\label{alg_1}
\KwIn{
A set of bags $\mathcal{D} = \{ \langle \mathcal{B}_i^+ \rangle \}_{i=1}^{p} \bigcup \{ \langle \mathcal{B}_i^- \rangle \}_{i=1}^{n}$.
}
\KwOut{
Instance labels  $y_{ij}$ in positive bags.
}
Initialize $w_{ij}=1$ \;
\While {not converged}
{
    Update $p(\scx_{ij} | y=1)$ and $p(\scx_{ij} | y=-1)$ using \eqref{eq_ekde}\;
    Update $w_{ij}$ using\eqref{eq_w_update}\;

}
Calculate voting score $f(\scx_{ij})$ for each instance using  \eqref{eq_votingscheme}\;

Return instances labels  $y_{ij}$ with \eqref{eq_sgn}\;
\end{algorithm}

To keep consistency with NegMin and CRANE that use $L_p$ distance, we adopt Gaussian kernel to measure the similarity, and restrict the covariance matrix to be isotropic, $\Sigma = \sigma^2 \text{I} $. We set different bandwidth values for positive class and negative class, and maximizing the overall class density difference to choose values from $\{0.001, 0.01,0.1, 1,  10, 100, 1000\}$. The instance labels are initialized with bag labels, i.e., $w_{ij}=1$. The algorithm is terminated when $w_{ij}$ is not changed, which only needs a few iterations in practice.

As for the convergence, a similar formulation where class conditional probability density and posterior probability are coupled is proposed in \cite{WangHMHQSD09}, and its closed-form solution is derived. In practice, they use an EM-style iterative method to avoid the expensive solution and have proven the convergence of the iterative process. Our eKDE can be considered as a variant of SSKDE in a weakly supervised scenario (their difference is analyzed in Section 3.5), therefore the convergence can be guaranteed. In our experiments, the algorithm usually terminates in a few iterations.

Computation cost: Although NegMin and CRANE do not use all of the negative instances to vote, they need to iterate through all of the instances to select the instances eligible to vote. Therefore, the computation cost of our voting scheme on the negative instances is theoretically identical to these negative mining methods. The computation from the instances in the positive bags increases our computation cost, making our method slower than the baselines.

\begin{table*}[!t]
\renewcommand{\arraystretch}{1.5}
\caption{Comparison of average precision.}
\label{tab_ap}
\centering
\begin{tabular*}{0.8\textwidth}{@{\extracolsep{\fill}}|l|c|c|c|c|c|c|c|c|c|c|c|c|}
\hline
\multirow{2}{*}{Methods} & \multirow{2}{*}{PittCar} & \multicolumn{11}{c|}{YTO} \\ \cline{3-13}
& & aero  & bird & boat & car & cat & cow & dog & hors & moto & trai & ave  \\ \hline
eKDE\_deep & 0.73 &  0.67 & 0.83 &  0.76 &  0.88 &  0.72 & 0.70 & 0.74 & 0.71 &  0.64 &  0.56 &   0.72 \\ \hline
eKDE & 0.49 &  0.56 & 0.83 &  0.59 &  0.75 &  0.61 & 0.70 & 0.59 & 0.35 &  0.27 &  0.17 &   0.54 \\ \hline
OBoW\cite{WangMHLZ17} & 0.47 &  0.22 & 0.50 & 0.35 &  0.88 & 0.56 & 0.65 &  0.56 & 0.41 & 0.27 & 0.33 & 0.47 \\ \hline
CRANE \cite{Tang13} & 0.21 &  0.00 &  0.33 & 0.35 & 0.75 & 0.33 & 0.45 &  0.26 &  0.24 & 0.27 & 0.17 & 0.32 \\ \hline
MILBoost \cite{Viola05} & 0.35 &  0.11 &  0.33 & 0.24 & 0.38 & 0.28 &  0.15 & 0.30 & 0.12 &  0.10 & 0.17 &  0.22 \\ \hline
\end{tabular*}
\end{table*}

\begin{figure*}[!t]
\centering
\begin{tabular}{ccccc}
  \vspace{+0.05in}
  eKDE & OBoW \cite{WangMHLZ17} & MILBoost \cite{Viola05} &  CRANE \cite{Tang13} \\ \vspace{+0.05in}
  {\includegraphics[width=0.23\linewidth]{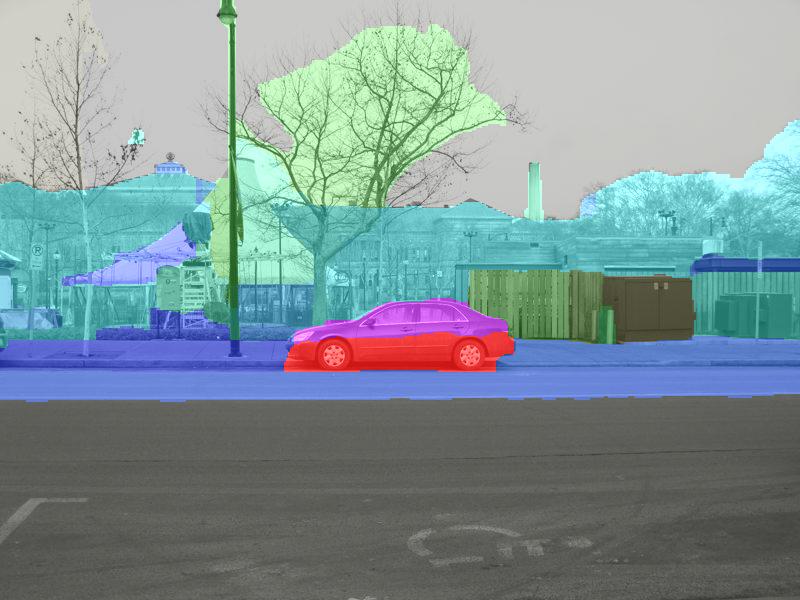}} &
  {\includegraphics[width=0.23\linewidth]{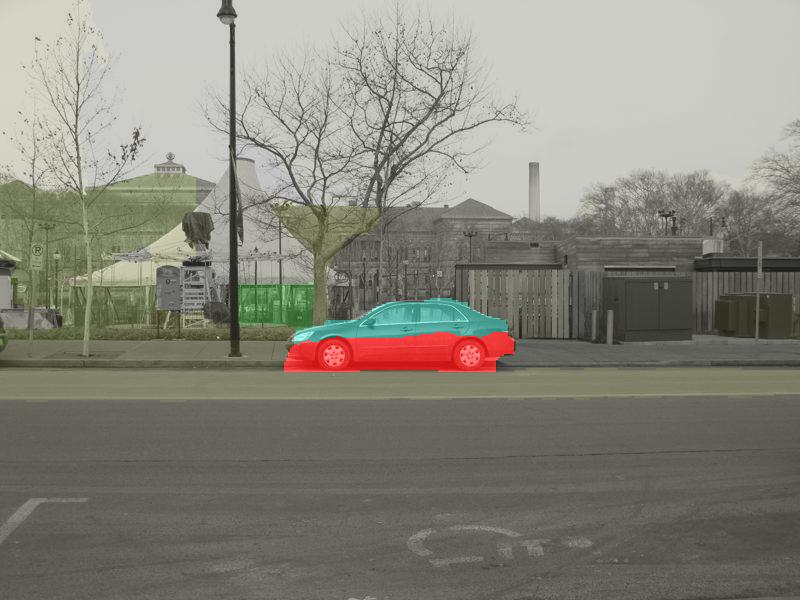}}   &
  {\includegraphics[width=0.23\linewidth]{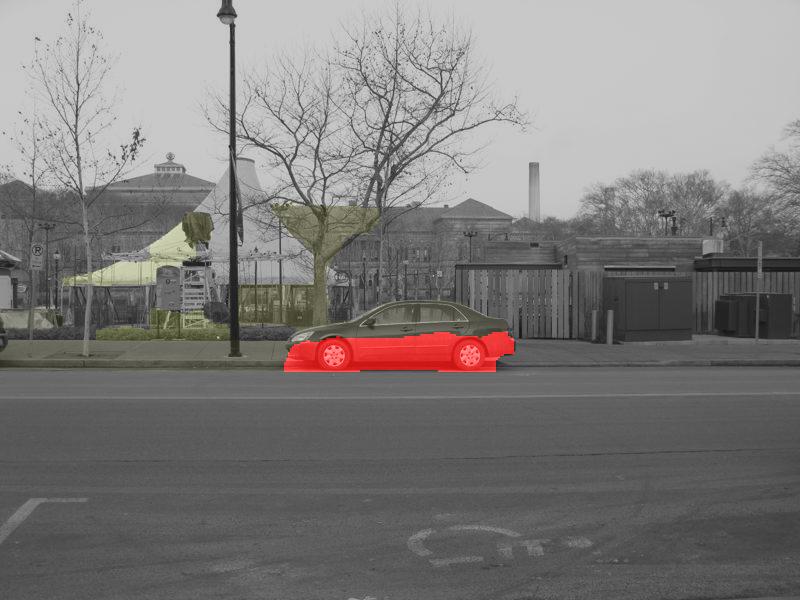}} &
  {\includegraphics[width=0.23\linewidth]{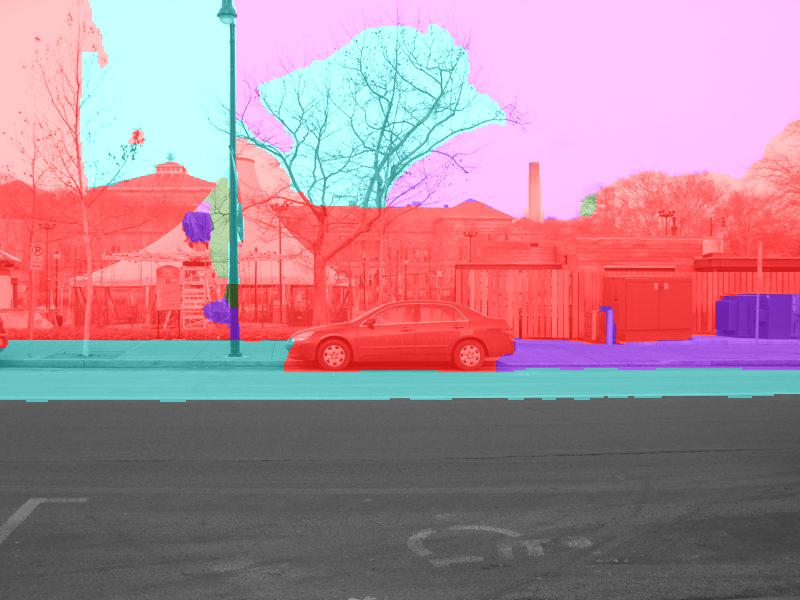}} \\ \vspace{+0.05in}
{\includegraphics[width=0.23\linewidth]{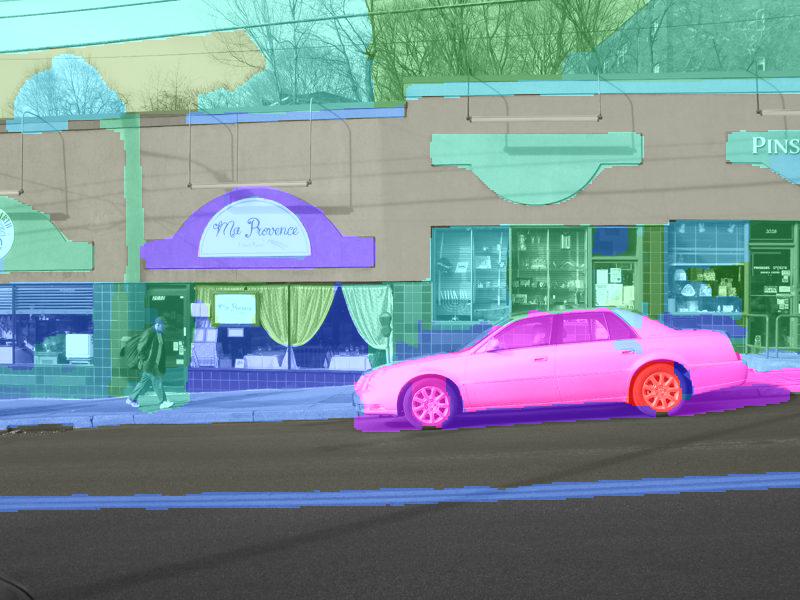}} &
  {\includegraphics[width=0.23\linewidth]{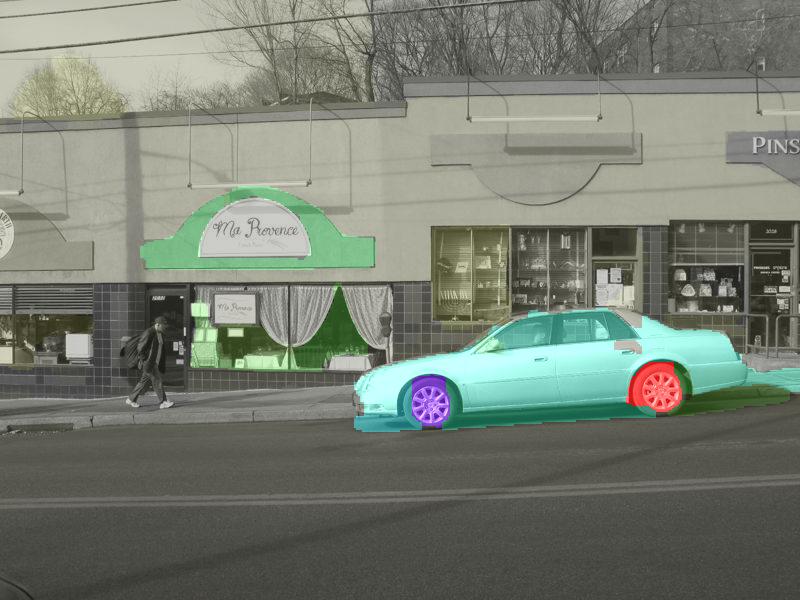}}   &
  {\includegraphics[width=0.23\linewidth]{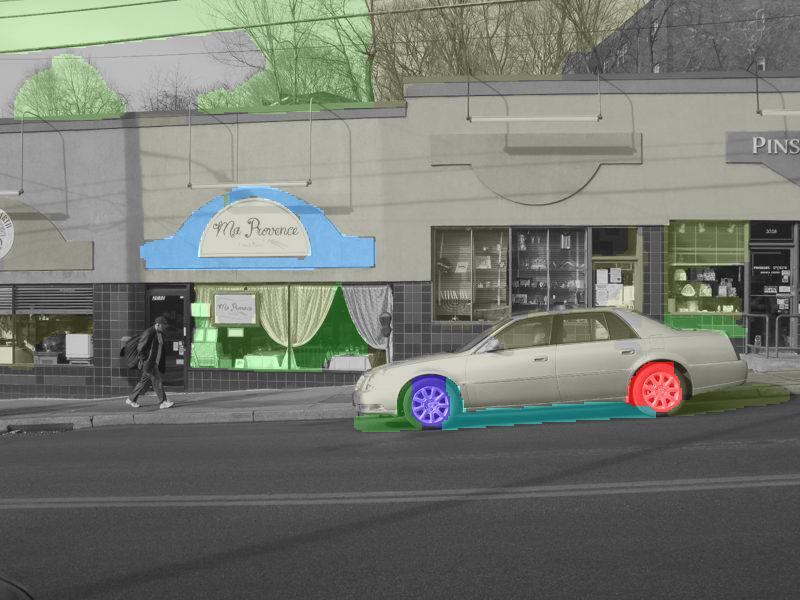}} &
  {\includegraphics[width=0.23\linewidth]{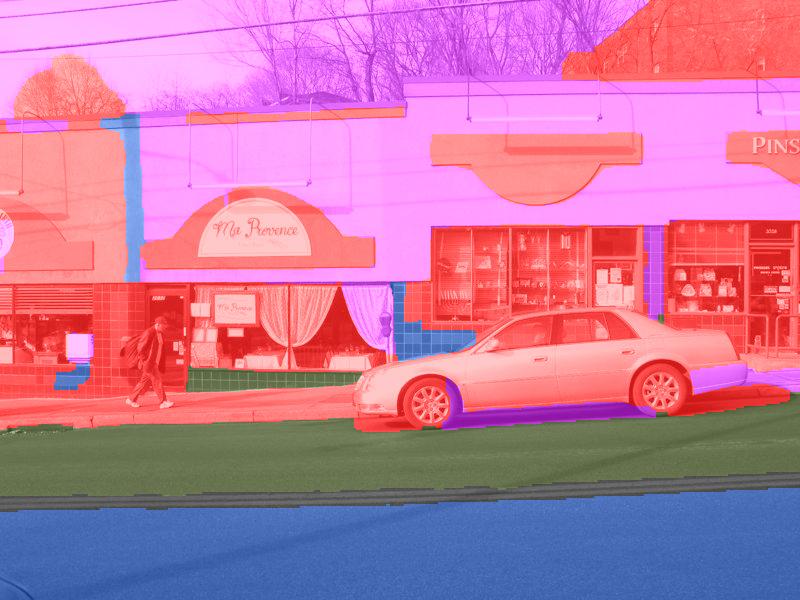}} \\ \vspace{+0.05in}
{\includegraphics[width=0.23\linewidth]{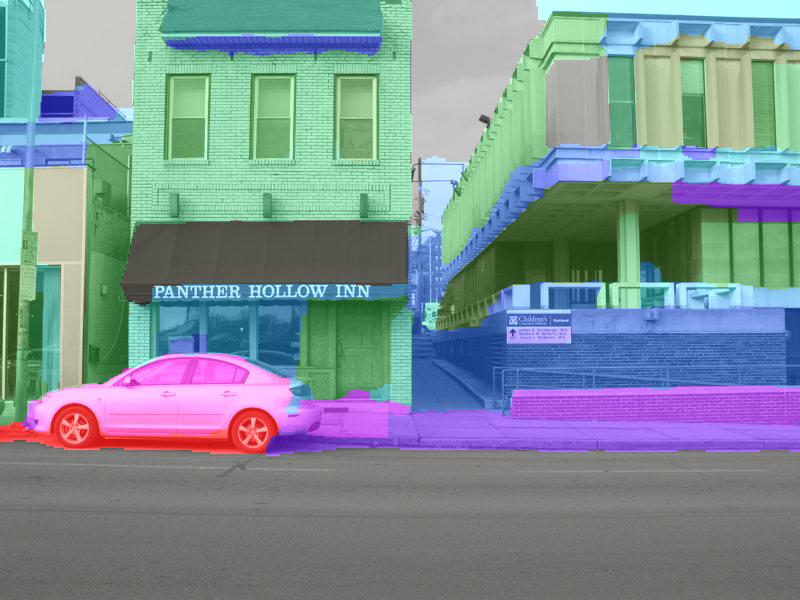}} &
  {\includegraphics[width=0.23\linewidth]{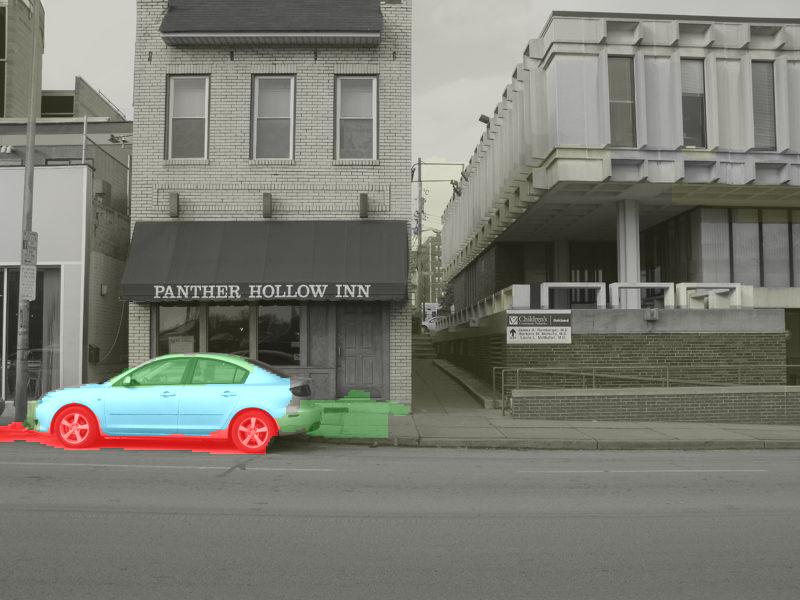}}   &
  {\includegraphics[width=0.23\linewidth]{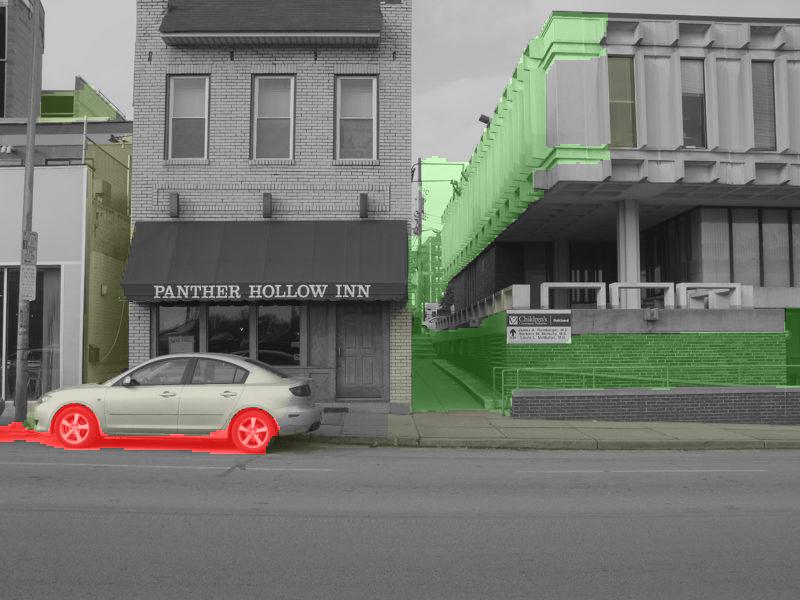}} &
  {\includegraphics[width=0.23\linewidth]{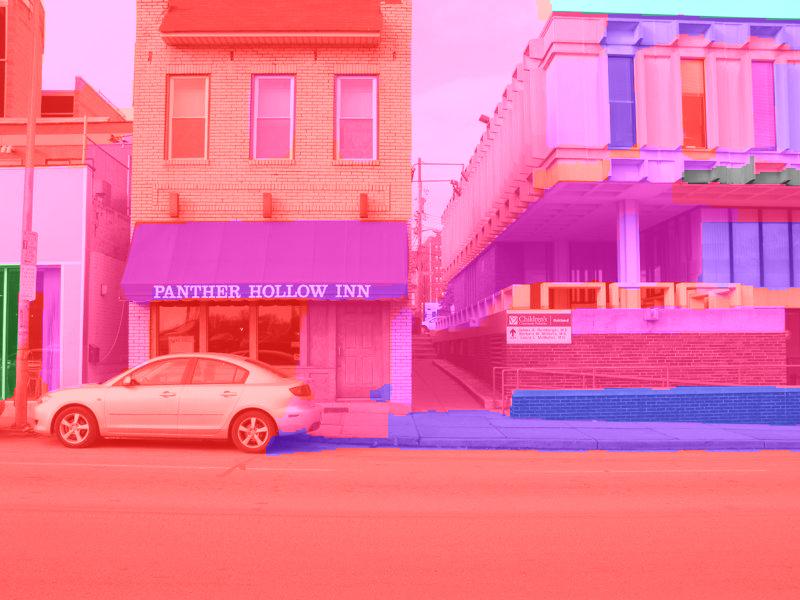}} \\ \vspace{+0.05in}
{\includegraphics[width=0.23\linewidth]{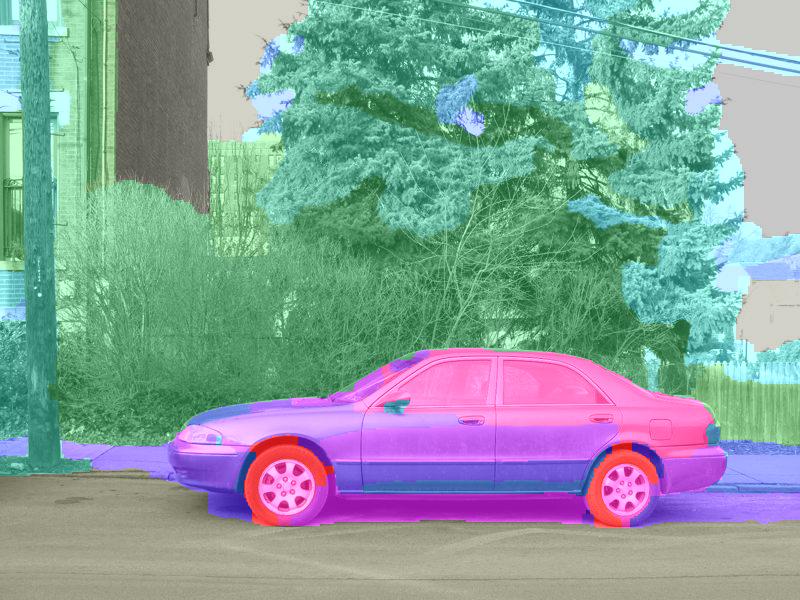}} &
  {\includegraphics[width=0.23\linewidth]{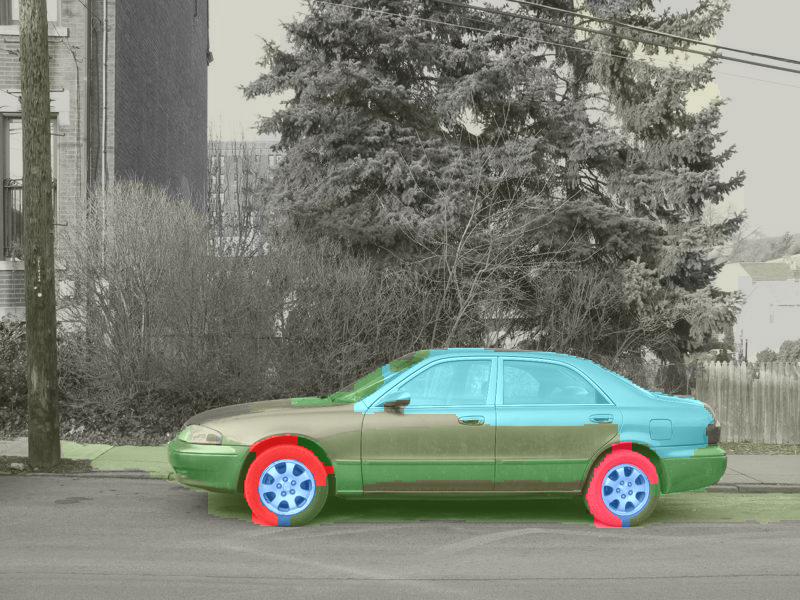}}   &
  {\includegraphics[width=0.23\linewidth]{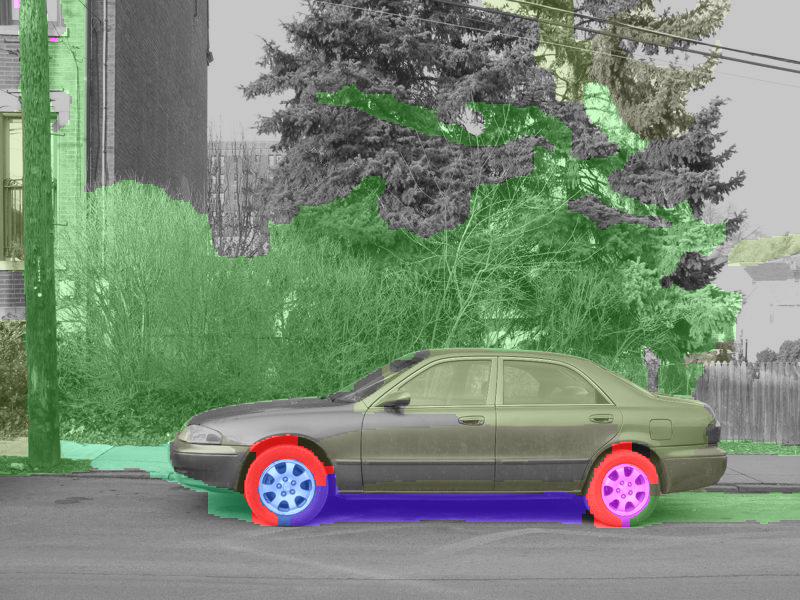}} &
  {\includegraphics[width=0.23\linewidth]{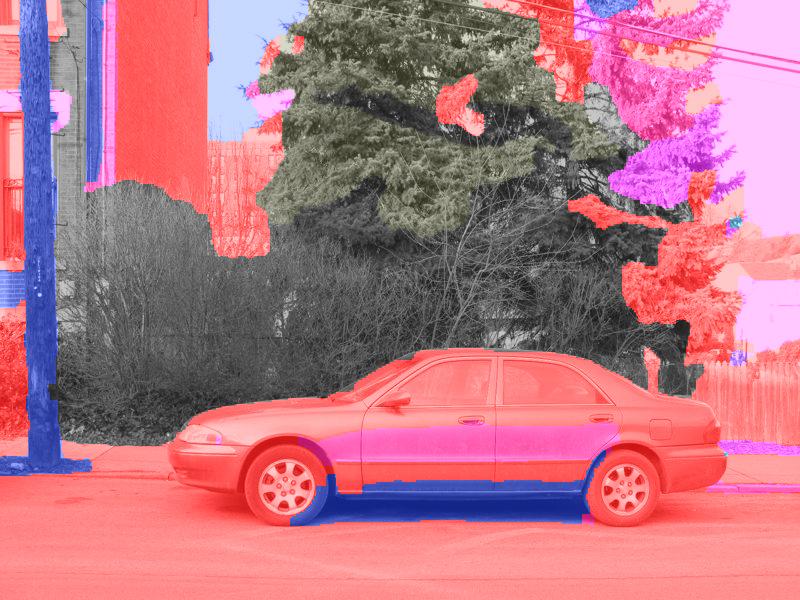}} \\ \vspace{+0.05in}
{\includegraphics[width=0.23\linewidth]{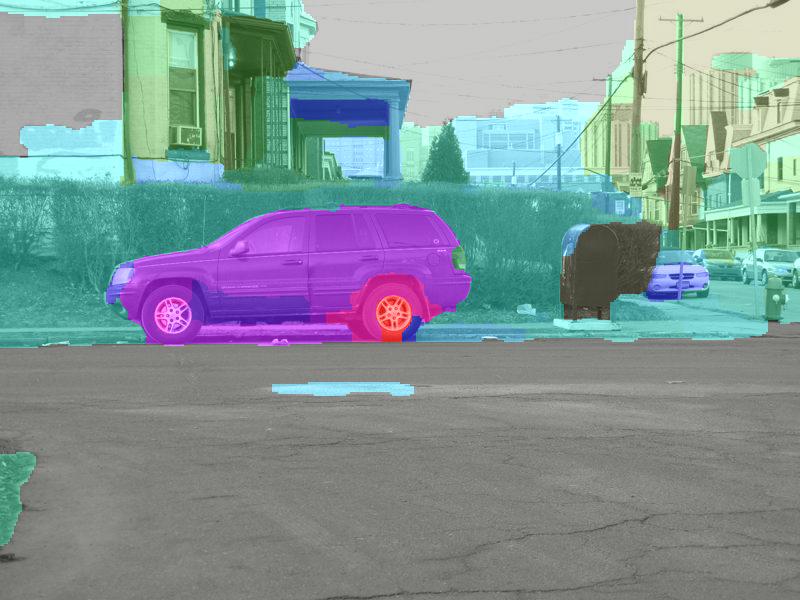}} &
  {\includegraphics[width=0.23\linewidth]{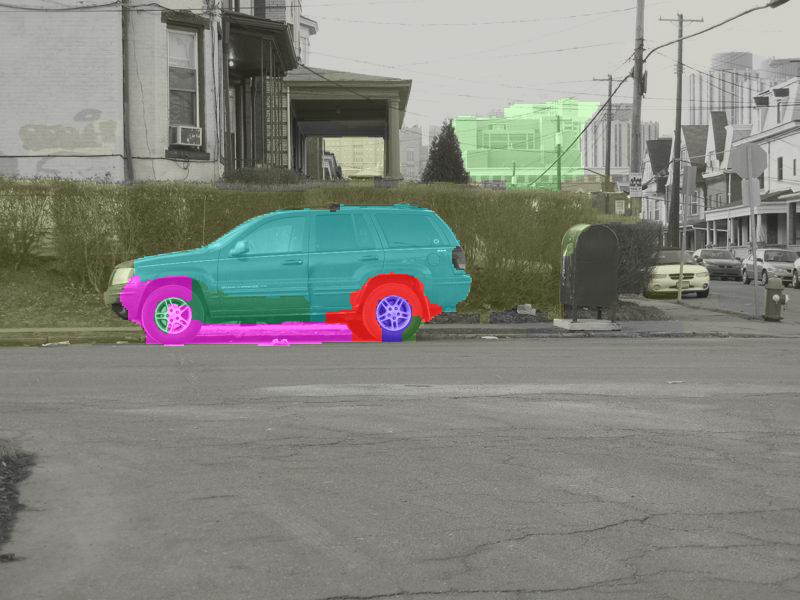}}   &
  {\includegraphics[width=0.23\linewidth]{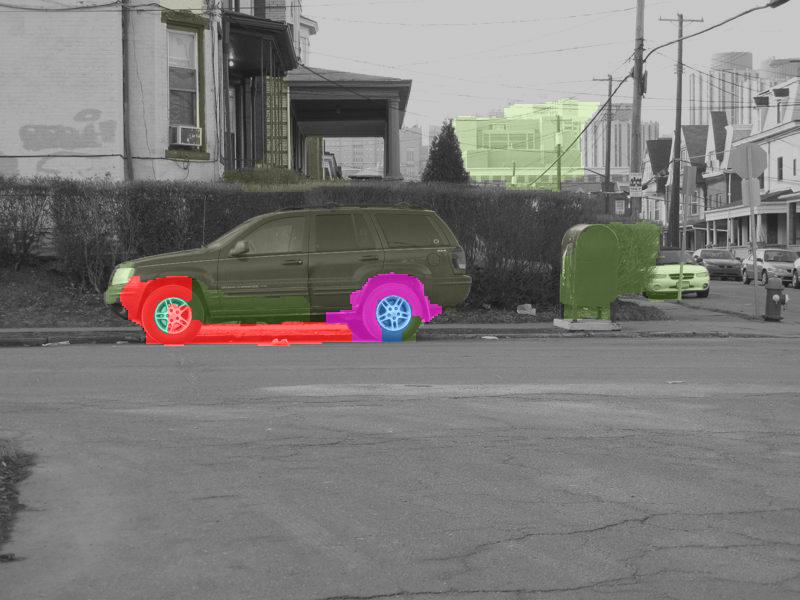}} &
  {\includegraphics[width=0.23\linewidth]{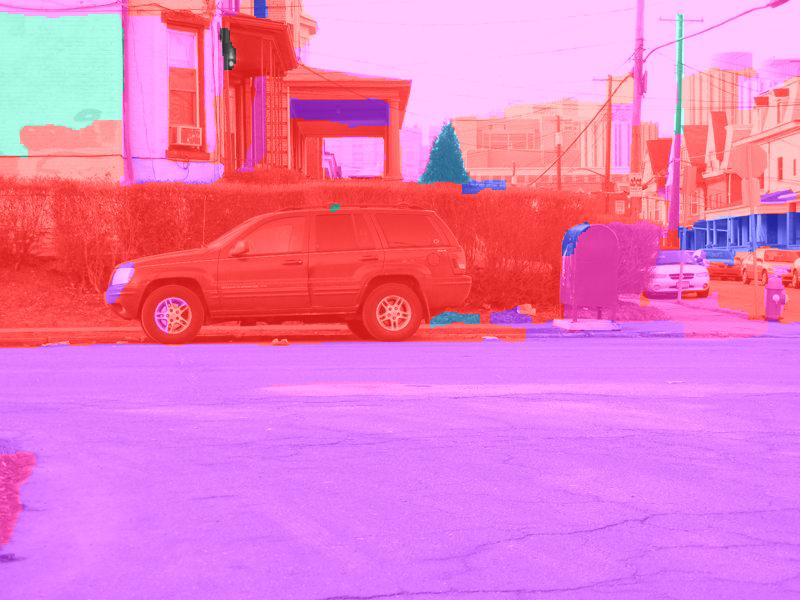}} \\
\end{tabular}
\caption{Visualization for some annotations. The warmer the color of the region is, the higher the likelihood of its belonging to an object. (Best viewed in color).}
\label{fig_vis}
\end{figure*}

\subsection{Difference from existing methods}
\label{sec_diff}
Given that our voting scheme has an interpretation from the eKDE perspective, we analyze its difference from SKDE \cite{DuWHY13} and SSKDE \cite{WangHMHQSD09}. By manually defining a conditional probability $p(\scx|\scx_{ij})$, SKDE obtains a density estimation p(\scx) from  observations with labels, and employs supervised mean shift to seek modes.
The main difference is that we try to estimate class-specific density $p(\scx|y)$ with weak labels, while SKDE aims at marginal density $p(\scx)$ under full supervision. This difference leads to SKDE cannot output category for instances, and it needs to try various starting points to seek different local maxima to obtain the key instances.

Our eKDE interpretation is also relative to SSKDE. It extends the conventional KDE to estimate the posterior probability $p(y|\scx)$ in a semi-supervised setting. In a semi-supervised setting, a little fraction of positive and negative instances are labelled to utilize a large amount of data that are totally unlabelled. While in a weakly supervised setting, a large amount of definite negative instances are available, and positive instances are given at bag level containing noises. This causes the difference in the way of using the labelled samples. For labelled instance, SSKDE calculates its posterior probability based partially on unlabelled set, whose relative importance is manually set by a parameter $t$ in \cite{WangHMHQSD09}. While in a weakly supervised setting, negative sample is large and their labels are definite, so their posterior probabilities do not rely on unlabelled sample. In addition, the weak labels provide good initialization for ambiguous instances to speed up the convergence.

\subsection{Second voting for refinement}
The above Algorithm \ref{alg_1} realizes weakly supervised segment annotation. It deals with each segment separately and cannot ensure the connectivity of the detected segments.  However, an object in an image/video must be a continues region. In other words, a set of adjacent regions form an object.
We therefore design a second round voting to integrate the fact and refine the annotation results. We first explore the adjacency of regions in an image/video, then for each region, the score is tuned with the mean score of its neighbours. Through this fine tuning, we encourage the adjacent regions to have similar voting scores, and expect  the region misclassified as background due to the similar appearance (e.g. a car window versus  house window when detecting a car) is expected to be corrected by location cues. In our experiments, such a refinement improves the results slightly.

\section{Experiments}
\label{sec_exp}
{\bf Setup and implementations.}
We compare our method with existing methods: OBoW \cite{WangMHLZ17}, CRANE \cite{Tang13}, NegMin\cite{Siva12}, MILBoost \cite{Viola05}.  Our weighted self-voting scheme is referred to as eKDE.

We consider Pittsburgh Car (PittCar) \cite{Nguyen09} and   YouTube-Objects (YTO) manually annotated in \cite{Tang13}.
PittCar dataset consists of $400$ images, where $200$ images contain cars. The background in the car images are similar street scenes to the non-car images, which is well suited to evaluate negative mining methods.
Some examples are shown in the first row of Fig. \ref{fig_data_example}. YTO dataset contains ten classes of videos collected from YouTube, see the last two rows of Fig. \ref{fig_data_example} for some examples. Tang et al. \cite{Tang13} generated a groundtruthed set by manually annotating the segments for 151 selected shots.

To keep consistency with NegMin and CRANE that use $L_p$ distance, we adopt Gaussian kernel, and restrict the covariance matrix to be isotropic, $\Sigma = \sigma^2 \text{I} $.
We use unsupervised methods \cite{Arbelaez11} and \cite{Xu12} to obtain  over-segmentation for images and videos respectively. We represent each segment using histogram of bag-of-visual-words obtained by mapping dense SIFT features \cite{Lowe04} into $1000$ words. For each description vector, we use L2 normalization before feeding to each model.
CRANE sweeps the threshold to generate precision/recall (PR) curves to conduct an evaluation.
In order to evaluate the discriminant performance of our method, we adopt the more popular evaluation metrics for object localization: the annotation is considered correct when the overlapping of the selected region and the ground-truth is larger than 0.5 for images and 0.125 for videos, then the average precision (the fraction of the correctly annotated images) is calculated. For fair comparison, we decide the threshold value for CRANE such that the number of the detected segments are the same as ours.

{\bf Results and analysis.}
We list the average precision in Table \ref{tab_ap}, where we can see that our method obtain better results than the baselines.

In order to analyze the above quantitative results, we visualize some annotation results in Fig. \ref{fig_vis}. As expected, our weighted voting method generates the best ranking of the segments¡¯ belonging to an object. MILBoost is usually able to locate the most discriminant region precisely, but the correctly annotated object regions are often too sparse, which leads to bad AP.
For CRANE, only negative instances that are nearby a segment could vote a penalty. This leads to many background regions in a positive image not penalized,  and these segments jointly have the identical maximum score 0.

{\bf Combined with deep features.}
Following \cite{ZhangMH17,WeiLWZ17,SalehASPAG18}, we leverage the DCNN models pre-trained for large scale classification task. We adopt the  VGG-NET \cite{SimonyanZ14a} pre-trained on the ImageNet in our method.
For the relatively simple PittCar dataset, we directly extract the feature maps using the original CNN parameters. For YTO dataset, we fine-tune the parameter before extracting features. Please note that we did not use the pixel-wise ground-truth during the tuning to ensure that our method is still weakly supervised.

For each image/frame, we resize it to $ 224\times 224$ and extract feature through the VGG model. The feature maps of  Conv5-4, Conv4-4, and Conv3-4 layers are collected, and are up-sampled to restore the original size. Then they are concatenated to a h*w*1280 3D tensor.
We then max-pool the vectors in a super-pixel to obtain a 1280-dimensional feature representation.
Our method using these deep features are referred to as eKDE\_deep. As shown in Table \ref{tab_ap}, replacing the SIFT feature by deep features in our voting can greatly improve the performance of segment annotation. This demonstrates that our algorithm can take advantage of deep CNN features and obtain much better results.
Note that we adopt different evaluation metrics from \cite{SalehASPAG18}, therefore higher values do not mean our method is better than theirs.
\section{Conclusion and discussion}
\label{sec_conc}
In this paper, we revisited the negative mining based methods under a voting framework. These methods can be considered as voting through only negative instances, which leads to their limitations: missing the useful information in positive bags and inability to determine the label of an instance. To overcome these limitations, we proposed a self-voting scheme involving the ambiguous instances as well as the definite negative ones. Each instance voted for its label with a weight computed from similarity. The ambiguous instances were assigned soft labels that were iteratively updated. We also derive an interpretation from eKDE and MAP, and analyzed the difference from the existing methods. In addition, deep CNN features can be included into the method to boost performance significantly. In future work, we will investigate how to construct end-to-end CNN for segment annotation.

\ifCLASSOPTIONcaptionsoff
  \newpage
\fi



\bibliographystyle{IEEEtran}
\bibliography{myref}
\end{document}